\pdfoutput=1

\documentclass[11pt]{article}

\usepackage[]{acl}

\usepackage{times}
\usepackage{latexsym}
\usepackage{amsmath}
\usepackage[T1]{fontenc}
\usepackage{algorithm}
\usepackage{algpseudocode}
\usepackage[utf8]{inputenc}

\usepackage{array}
\usepackage{booktabs}
\usepackage{microtype}
\usepackage{multirow}
\usepackage{inconsolata}
\usepackage{multicol} 
\usepackage{graphicx}
\usepackage{comment}
\usepackage{enumitem}
\usepackage{relsize}
\usepackage{booktabs}
\usepackage{tabularx}
\usepackage{multirow}
\usepackage{xspace}
\usepackage{xcolor}
\usepackage{subcaption}
\usepackage{hyperref}

\usepackage{amssymb}
\usepackage{amsthm}
\usepackage{amsfonts}
\usepackage{mathtools}

\usepackage{cleveref}

\usepackage{tikz}
\usetikzlibrary{arrows.meta}
\usetikzlibrary{shapes.arrows}
\usetikzlibrary{shapes.misc}
\usetikzlibrary{fit}
\usetikzlibrary{calc,shapes}
\usetikzlibrary{tikzmark}
\usetikzlibrary{arrows}

\usepackage{xfrac}
\crefname{chapter}{Chapter}{}
\crefname{section}{Section}{Sections}
\crefformat{section}{Section~#2#1#3}

\crefname{table}{Table}{Tables}
\crefname{figure}{Figure}{Figures}
\crefname{algorithm}{Alg.}{Algs.}
\crefname{line}{Line}{Lines}
\crefname{appendix}{App.}{}
\crefname{chapter}{Chapter}{Chapters}

\crefname{thm}{Theorem}{Theorems}
\crefname{prop}{Proposition}{Propositions}
\crefname{definition}{Definition}{Definitions}
\crefname{lemma}{Lemma}{Lemmas}
\crefname{cor}{Corollary}{Corollaries}
\crefname{equation}{Eq.}{Eqs.}

\creflabelformat{equation}{#2\textup{#1}#3}
\crefrangelabelformat{equation}{(#3#1#4--#5#2#6)}

\theoremstyle{definition}

\newcolumntype{C}{>{\centering\arraybackslash}X}

\renewcommand{\ldots}{\ensuremath{{\ldotp\kern-0.2em\ldotp\kern-0.2em\ldotp}}}

\renewcommand{\cdots}{\ensuremath{{\cdotp\kern-0.2em\cdotp\kern-0.2em\cdotp}}}
\renewcommand{\dots}{\ensuremath{{\ldotp\kern-0.2em\ldotp\kern-0.2em\ldotp}}}

\newcommand{\defn}[1]{\textbf{#1}}

\renewcommand{\hat}[1]{\widehat{#1}}
\renewcommand{\bar}[1]{\overline{#1}}

\newcommand{\defeq}[0]{\mathrel{\stackrel{\textnormal{\tiny def}}{=}}}

\newcommand{\checkNotation}[1]{{#1}} 

\newcommand{\checkNotationOp}[3]{\checkNotation{#1 {\normalcolor {#2}} #3}}

\DeclareMathOperator*{\minCall}{min\,}

\renewcommand{\min}[1]{\checkNotation{\minCall\limits_{\mathclap{\substack{{\normalcolor #1}}}}\,}}

\newcommand{\abs}[1]{\checkNotationOp{\left\lvert}{#1}{\right\rvert}}

\newcommand{\randAlg}{\checkNotation{A}}
\newcommand{\privacyBudget}{\checkNotation{\epsilon}}
\newcommand{\errorProb}{\checkNotation{\delta}}
\newcommand{\dataset}{\checkNotation{D}}
\newcommand{\datasetAlt}{\checkNotation{D'}}
\newcommand{\outputs}{\checkNotation{\mathcal{S}}}
\newcommand{\prob}[1]{\Pr\left[ #1 \right]}
\newcommand{\dPr}{$(\privacyBudget, \errorProb)$-DP}
\newcommand{\batch}{\checkNotation{L}}

\newcommand{\clippingBound}{\checkNotation{C}}
\newcommand{\learningRate}{\checkNotation{\gamma}}
\newcommand{\noise}[1]{\checkNotation{\mathcal{N} \!\left( #1 \right)}}

\newcommand{\betaA}{\checkNotation{\beta_1}}
\newcommand{\betaB}{\checkNotation{\beta_2}}
\newcommand{\gradient}[1]{\checkNotation{g(#1)}}
\newcommand{\gradientNoise}[1]{\checkNotation{\tilde{g}(#1)}}

\newcommand{\sensitivitySolo}[1]{\checkNotation{S_{#1}}}
\newcommand{\sens}[1]{\checkNotation{\bar{S}_{#1}}}
\newcommand{\uncer}[1]{\checkNotation{\bar{U}_{#1}}}
\newcommand{\importance}[1]{\checkNotation{I_{#1}}}
\newcommand{\impSmooth}[1]{\checkNotation{\hat{I}_{#1}}}
\newcommand{\impNorm}[1]{\checkNotation{\bar{I}_{#1}}}
\newcommand{\median}[1]{\checkNotation{\mathrm{median}\!\left(#1\right)}}
\newcommand{\mean}[1]{\checkNotation{\mathrm{mean}\!\left(#1\right)}}

\newcommand{\weights}[1]{\checkNotation{\omega_{#1}}}

\newcommand{\adanoise}{\checkNotation{\textsc{AnaDP}}}

\usepackage{todonotes}
\makeatletter
\newcommand*\iftodonotes{\if@todonotes@disabled\expandafter\@secondoftwo\else\expandafter\@firstoftwo\fi}  
\makeatother

\title{Fine-Tuning Language Models with Differential Privacy through \\Adaptive Noise Allocation}

\author{Xianzhi Li\textsuperscript{1}, Ran Zmigrod\textsuperscript{2}, {\bf Zhiqiang Ma\textsuperscript{2}}, {\bf Xiaomo Liu\textsuperscript{2}}, Xiaodan Zhu\textsuperscript{1}\\
\textsuperscript{1}Department of Electrical and Computer Engineering \& Ingenuity Labs Research Institute \\ Queen's University\\
\textsuperscript{2}J.P. Morgan AI Research\\
\small{\{\texttt{li.xianzhi, xiaodan.zhu}}\}\text{\texttt{@queensu.ca}} \\\small{\{\texttt{ran.zmigrod, zhiqiang.ma, xiaomo.liu}}\}\text{\texttt{@jpmchase.com}}
\\ }

\begin{document}
\maketitle
\begin{abstract}
Language models are capable of memorizing detailed patterns and information, leading to a double-edged effect: they achieve impressive modeling performance on downstream tasks with the stored knowledge but also raise significant privacy concerns. Traditional differential privacy based training approaches offer robust safeguards by employing a uniform noise distribution across all parameters. 
However, this overlooks the distinct sensitivities and contributions of individual parameters in privacy protection and often results in suboptimal models.
To address these limitations, we propose \adanoise{}, a novel algorithm that adaptively allocates additive noise based on the importance of model parameters.
We demonstrate that \adanoise{} narrows the performance gap between regular fine-tuning and traditional DP-SGD based fine-tuning on a series of datasets while maintaining the required privacy constraints.
\end{abstract}

\section{Introduction}

Language models have achieved remarkable success and shown impressive abilities in a wide range of tasks  \cite{almazrouei2023falcon, touvron2023llama, team2023gemini}. Their advanced capabilities in memorizing detailed information and patterns in data as well as making connections among them have not only helped language models to achieve impressive modeling performance on downstream tasks, but also raised significant and ubiquitous privacy concerns if it is not properly handled~\cite{neel2023privacy, mireshghallah2023can, yao2024survey}.

Differential Privacy (DP) is a principled framework for mitigating privacy risks, providing theoretical guarantees that prevent inferring the presence or absence of an individual's data in a model's output~\citep{abadi2016deep, dwork2006differential}. Conventional DP-enhanced fine-tuning offers robust safeguards by assuming a uniform noise distribution across all parameters to protect privacy~\citep{kerrigan2020differentially,yu2021large,li2021large}. Recent work has begun exploring DP in Parameter Efficient Fine Tuning (PEFT)~\citep{yu2021differentially, bu2022differentially}, which is built on the same assumption on the additional tunable parameters. Unfortunately, such an assumption overlooks the distinct sensitivities and contributions of individual parameters in privacy protection and often results in suboptimal models.


In this paper, we introduce \adanoise{}, a novel DP method that adaptively distributes the noise and privacy budget among a language model's parameters during fine-tuning, based on their importance to the model at a given training step.
Our work was inspired by \citet{zhang2023adaptive}, who utilized the sensitivity and uncertainty of parameters for model pruning.
Importantly, our approach not only respects the inherent heterogeneity of parameter significance but also maintains strong privacy protection.
The proposed integration addresses the key challenges in effectively measuring the contributions of parameters and ensures that models are trained stably. 
We demonstrate that \adanoise{} consistently improves performance over the traditional DP fine-tuning under the same privacy budget and bridges the gap between traditional DP and non-DP fine-tuning (no privacy guarantee).
The contributions of our work are summarized below:
\begin{itemize}[left=1.5mm, itemsep=0mm]
\item 
We propose \adanoise{}, a novel algorithm for fine-tuning language models while maintaining privacy guarantees.
To the best of our knowledge, this is the first DP method that distributes the privacy budget based on Transformer parameters' importance non-uniformly.

\item 
We empirically demonstrate that \adanoise{} outperforms the standard DP approaches on the Glue benchmark \citep{wang2018glue}
in multiple training paradigms (e.g. both full fine-tuning and PEFT). 
\item
We conduct further analysis on privacy exposure risk and find that \adanoise{} offers the same robust privacy protection as the conventional DP method.


    
\end{itemize}

\section{Related Work}

\paragraph{Differential Privacy.}
DP is a principled approach to ensuring privacy. The concept of DP was formalized by \citet{dwork2006differential}, who introduced the definition and foundational mechanisms of DP. In machine learning, \citet{abadi2016deep} introduced the widely used Differentially Private Stochastic Gradient Descent (DP-SGD). Adaptive Differential Privacy is a recent development. Research \cite{gong2020preserving, chen2023differentially}  have been developed to preserve adaptive DP in deep neural networks. 
However, these methods fall short in capturing the complex parameter interactions within transformers, potentially leading to suboptimal models and trade-offs between privacy and utility.

\paragraph{Fine-Tuning and PEFT.}
Full fine-tuning used to be a prominent approach but can be resource-intensive and less efficient \citep{lester2021power, tay2022efficient}. PEFT has emerged as another option for effectively training LLMs. Many PEFT techniques, such as LoRA \cite{hu2021lora}, Adapters \cite{houlsby2019parameter}, and prefix tuning \cite{li2021prefix} have been proposed to tune small, additional modules instead of the whole model. \citet{he2021towards} provided a unified view revealing the connections among various parameter-efficient transfer learning methods. Recent work by \cite{zhang2023adaptive} introduced AdaLoRA to dynamically adjust the amount of parameter tuning based on the task and model requirements. Despite their efficiency, integrating these methods with privacy-preserving techniques remains an area that requires further exploration.

\section{The \adanoise~Model}
The integration of Differential Privacy for language model fine-tuning is crucial for deploying LLMs in privacy-sensitive applications. In this work, we introduce \adanoise{}, an adaptive noise allocation DP training method based on the importance score of models' parameters, which provides a generic solution that can be applied to a wide range of LLMs.  
The fundamental idea is that adding less noise to the parameters that are more important and more to the less important parameters would help improve the model's utility given the same privacy budget.
This section describes the construction and correctness of \adanoise{}, whose pseudocode is given in \cref{alg:adanoise}.


\definecolor{commentcolor}{rgb}{0.4, 0.22, 0.33}

\newcommand{\rightcomment}[1]{{\color{commentcolor} \(\triangleright\) {\footnotesize\textit{#1}}}}
\algrenewcommand{\algorithmiccomment}[1]{\hfill \rightcomment{#1}}  
\algnewcommand{\LineComment}[1]{\State \rightcomment{#1}}
\algrenewcommand\algorithmicindent{0.7em}%

\begin{algorithm}[t]
\caption{\adanoise{} Algorithm}
\label{alg:adanoise}
\begin{algorithmic}[1]
\State \textbf{Input:} Training batches $\mathcal{L}=\{\batch_1, \dots, \batch_T\}$, Initial parameter weighs $\weights{0}$, noise multiplier $\sigma_0$
\State \textbf{Hyper-parameters:} $\alpha$, $\betaA$, $\betaB$, clipping threshold $\clippingBound$, learning rate $\learningRate$
\State $\sensitivitySolo{0}\gets\boldsymbol{0}$, $\sens{0}\gets\boldsymbol{0}$, $\uncer{0}\gets\boldsymbol{0}$
\For{$\batch_t \in \mathcal{L}$}
    \State Compute gradients $\gradient{\batch_t}$ \hspace{0.1cm} 
    \State $\sensitivitySolo{t} \gets \abs{\gradient{\batch_t} \cdot \weights{t-1}} $ \Comment{Compute Sensitivity}
    \State $\sens{t} \gets \betaA \sens{t-1} + (1 - \betaA) \sensitivitySolo{t}$ \Comment{\cref{eq:sens-avg}}
    \State $\uncer{t} \gets \betaB \uncer{t-1} + (1 - \betaB) \abs{\sens{t} - \sensitivitySolo{t}}$ \Comment{\cref{eq:uncer-avg}}
    \State $\importance{t} \gets \sens{t} \cdot \uncer{t}$ \Comment{\cref{eq:imp}}
    \State $\mu \gets \mean{\frac{\importance{t} - \median{\importance{t}}}{q_{1}(\importance{t}) - q_{2}(\importance{t})}}$ \Comment{Mean importance}
    \State $\impSmooth{t} \gets (1-\alpha) \!\! \left( \frac{\importance{t} - \median{\importance{t}}}{q_{1}(\importance{t}) - q_{2}(\importance{t})} \right) \!\! + \alpha \mu$ \Comment{\cref{eq:imp-smooth}}
    \State $\impNorm{t} \gets \impSmooth{t} - \left(\mean{\impSmooth{t}} - 1\right)$ \Comment{\cref{eq:imp-norm}}
    \State $\gradientNoise{\batch} \gets \min{}\!\left(\gradient{\batch}, \clippingBound\right) + \noise{\frac{\sigma_0^2}{\impNorm{t}}}$ \Comment{\cref{eq:grad-adanoise}}
    \State $\weights{t} \gets \weights{t-1} - \learningRate \gradient{\batch_t}$ \Comment{Update weights}
\EndFor
\State \textbf{Output:} Updated parameters $\weights{T}$
\end{algorithmic}
\end{algorithm}

We follow \citet{dwork2006differential}'s definition of DP.
Specifically, we achieve DP through a \defn{randomized algorithm} $\randAlg$ over an output space $\outputs$.
Given a privacy budget $\privacyBudget$ and error probability $\errorProb$, we say $\randAlg$ is \defn{($\boldsymbol{\privacyBudget}$, $\boldsymbol{\errorProb}$)-differentially private} (\defn{($\boldsymbol{\privacyBudget}$, $\boldsymbol{\errorProb}$)-DP}) if for any neighboring datasets $\dataset$ and $\datasetAlt$, which differ in exactly one data record, the following inequality holds:
\begin{equation}
    \prob{\randAlg(\dataset)\in\outputs} \leq e^{\privacyBudget} \prob{\randAlg(\datasetAlt)\in\outputs} + \errorProb
\end{equation}



\noindent where privacy budget $\privacyBudget$ is a measure of the amount of privacy loss allowed during training. Past methods for achieving \dPr~typically add a uniform Gaussian noise to the parameters.
More formally, given a batch $\batch$, we can define adding Gaussian noise to the model's gradient, $\gradient{\batch}$ as:
\begin{equation}\label{eq:grad-noise}
    \gradientNoise{\batch} \defeq \min{}\!(\gradient{\batch}, C) + \noise{C\sigma^2}
\end{equation}
where $C$ is a clipping threshold and $\noise{C\sigma^2}$ is Gaussian noise with mean $0$ and variance $C\sigma^2$.
$C$ and $\sigma$ are fixed and computed based on the privacy budget \citep{abadi2016deep}.
In this work, we explore a tunable $\sigma$ to realize a better trade-off between privacy and utility.
We aim to tailor the noise distribution across different parameters and the key objective is to determine the importance of each parameter.


\begin{table*}[ht]
\centering
\small
\begin{tabular}{cc ccccc ccccc}
\toprule
 \multicolumn{2}{c}{ \bf Method} & \multicolumn{5}{c}{\bf Roberta-base} & \multicolumn{5}{c}{\bf Roberta-large} \\
 \cmidrule(lr){0-1}\cmidrule(lr){3-7} \cmidrule(lr){8-12}
 
 \bf Paradigm & \bf Privacy Alg. & \bf SST-2 & \bf QNLI & \bf MNLI & \bf QQP & \bf Avg. & \bf SST-2 & \bf QNLI & \bf MNLI & \bf QQP & \bf Avg. \\ 
\midrule
\multirow{3}{*}{Full} & w/o DP  & 94.8 & 92.8 & 87.6 & 91.9 & 91.8 & 96.4 & 94.7 & 90.2 & 92.2 & 93.4  \\ 
 & DP & 91.5 & 87.9 & 83.4 & 86.4 & 87.5 & 95.0 & 91.2 & 87.2 & 86.8 & 90.1 \\ 
  & \adanoise & 92.5 & 89.1 & 84.0 & 87.6 & 88.3 & 95.2 & 92.3 & 87.9 & 88.5 & 91.0 \\ 
 \midrule
 \multirow{3}{*}{LoRA} & w/o DP & 95.1 & 93.3 & 87.5 & 90.8 & 91.7 & 96.2 & 94.9 & 90.6 & 91.6 & 93.3 \\ 
 & DP & 92.2 & 87.3 & 83.5 & 85.7 & 87.2 & 95.3 & 90.8 & 87.8 & 87.4 & 90.3 \\ 
 & \adanoise & 93.4 & 88.8 & 83.9 & 86.5 & 88.2 & 95.7 & 91.9 & 88.1 & 87.7 & 91.0 \\ 
 \midrule
 \multirow{3}{*}{Adapter} & w/o DP & 94.7 & 93.0 & 87.3 & 90.6 & 91.4 & 96.4 & 94.7 & 90.3 & 91.5 & 93.2 \\ 
  & DP & 92.5 & 87.5 & 83.4 & 85.6 & 87.3 & 93.9 & 90.7 & 87.7 & 86.3 & 89.7 \\ 
  & \adanoise & 93.4 & 88.0 & 84.4 & 86.3 & 88.0 & 94.8 & 91.8 & 88.7 & 87.7 & 90.8 \\  
  \bottomrule
\end{tabular}
\caption{Performance Comparison (accuracy) for \adanoise{} with baselines using full fine-tuning, LoRA, and Adapter tuning.
The performance differences between \adanoise{} and DP are all statistically significant with p < 0.05 under the one-tailed paired t-test.}
\label{tab:performance_metrics}
\end{table*}

\paragraph{Parameter Importance.}
In order to gauge parameters' importance, our work is inspired by \citet{zhang2023adaptive}, which calculates importance based on the sensitivity and uncertainty of the parameter for model pruning.
We use the moving averages of the sensitivity and uncertainty of the model parameters at training step $t$:
\begin{align}
    \sens{t} &\defeq \beta_1 \sens{t-1} + (1 - \beta_1) \sensitivitySolo{t} \label{eq:sens-avg} \\
    \uncer{t} &\defeq \beta_2 \uncer{t-1} + (1 - \beta_2) \abs{\sens{t} - \sensitivitySolo{t}} \label{eq:uncer-avg}
\end{align}
where $\beta_1,\beta_2\in[0,1]$ are hyper-parameters to control the move average rate.
Additionally, $\sensitivitySolo{t}\defeq\abs{\gradient{\batch_t}\cdot\weights{t-1}}$ is the \defn{sensitivity} of the model weights at step $t$.\footnote{The \defn{uncertainty} of a parameter at step $t$ is the absolute difference between its sensitivity at step $t$ and its moving average $\sens{t}$.}
The \defn{importance metric} is then the element-wise product of the sensitivity and uncertainty
\begin{equation}\label{eq:imp}
    \importance{t} \defeq \sens{t} \cdot \uncer{t}.
\end{equation}
This formulation ensures that parameters with moderate sensitivity but high uncertainty are still considered important, which prevents prematurely discarding parameters that could become important as training progresses.

\paragraph{Importance  Normalization.}\label{sec:methods_normalization}
Using $\importance{t}$ as defined in \cref{eq:imp} may lead to zero-gradients.
Therefore, we smooth the importance scores
\begin{equation}\label{eq:imp-smooth}
    \impSmooth{t} \defeq (1-\alpha) \!\! \left( \frac{\importance{t} - \median{\importance{t}}}{q_{1}(\importance{t}) - q_{2}(\importance{t})} \right) \!\! + \alpha \mu
\end{equation}
where $\alpha\in[0,1]$ is a smoothing parameter, $q_{1}(\importance{t})$ and $q_{2}(\importance{t})$ are chosen quantiles of $\importance{t}$, and $\mu$ is the mean of the scaled normalized vector.

$\impSmooth{t}$ gives a scaled distribution of importance across the model's parameters.
In order to ensure \dPr{}, we further adjust distribution to be centered at one
\begin{equation}\label{eq:imp-norm}
    \impNorm{t} \defeq \impSmooth{t} - \left(\mean{\impSmooth{t}} -1\right).
\end{equation}
This means that the overall noise added to the model will follow the same distribution as that of \citet{abadi2016deep} who uses a uniform noise across all parameters.
As the overall noise added is the same, \adanoise{} satisfies the \dPr{} guarantees following the proof of \citet{abadi2016deep}.
The smoothed importance score is utilized to adaptively add noise to the gradient.
Replicating \cref{eq:grad-noise}, our new gradient is:
\begin{equation}\label{eq:grad-adanoise}
\gradientNoise{\batch} = \min{}\!(\gradient{\batch}, C) + \noise{\frac{\sigma_0^2}{\impNorm{t}}}
\end{equation}
where $\sigma_0$ is a noise multiplier that achieves \dPr{} and is selected following \citet{abadi2016deep}.




\section{Experiments}

\subsection{Experimental Setup}
We evaluate the performance of \adanoise{} against the traditional DP-SGD method \citep{abadi2016deep}, DP-PEFT method~\citep{yu2021differentially} as well as regular fine-tuning (i.e., no privacy guarantees).
Same as in previous work~\citep{wu2023depn, yu2021differentially}, we run our three privacy configurations using RoBERTa (base and large) \citep{liu2019roberta} in the full fine-tuning setting as well as on two state-of-the-art PEFT methods: LoRA \cite{hu2021lora} and Adapter \cite{houlsby2019parameter}.\footnote{More details including hyperparameters chosen are given in Appendix \ref{sec:appendix_experiment}.}
Each of the combinations above is evaluated against four datasets from the Glue benchmark~\citep{wang2018glue} which is used in past work to evaluate conventional DP-SGD: SST-2~\citep{socher-etal-2013-recursive}, QNLI~\citep{rajpurkar2016squad}, MNLI~\citep{williams-etal-2018-broad}, and QQP.\footnote{https://data.quora.com/First-Quora-Dataset-Release-Question-Pairs}

We further conduct privacy protection experiments.
Our experiments follow those of \cite{wu2023depn}; we train a model using contexts containing private information (the Enron email dataset~\cite{klimt2004enron}), and then compute the leakage risk of the privacy information. The Enron email dataset is comprised of over $500,\!000$ emails that contain sensitive information such as person names and phone numbers
Specifically, we use the Mean Reciprocal Rank (MRR) for person name and exposure \cite{carlini2019secret} metric for telephone numbers to show the risk of privacy leakage. 




\subsection{Accuracy Results}\label{sec:accuracy}
\begin{figure*}[t]
    \centering
    \resizebox{0.8\linewidth}{!}{\includegraphics[width=\linewidth]{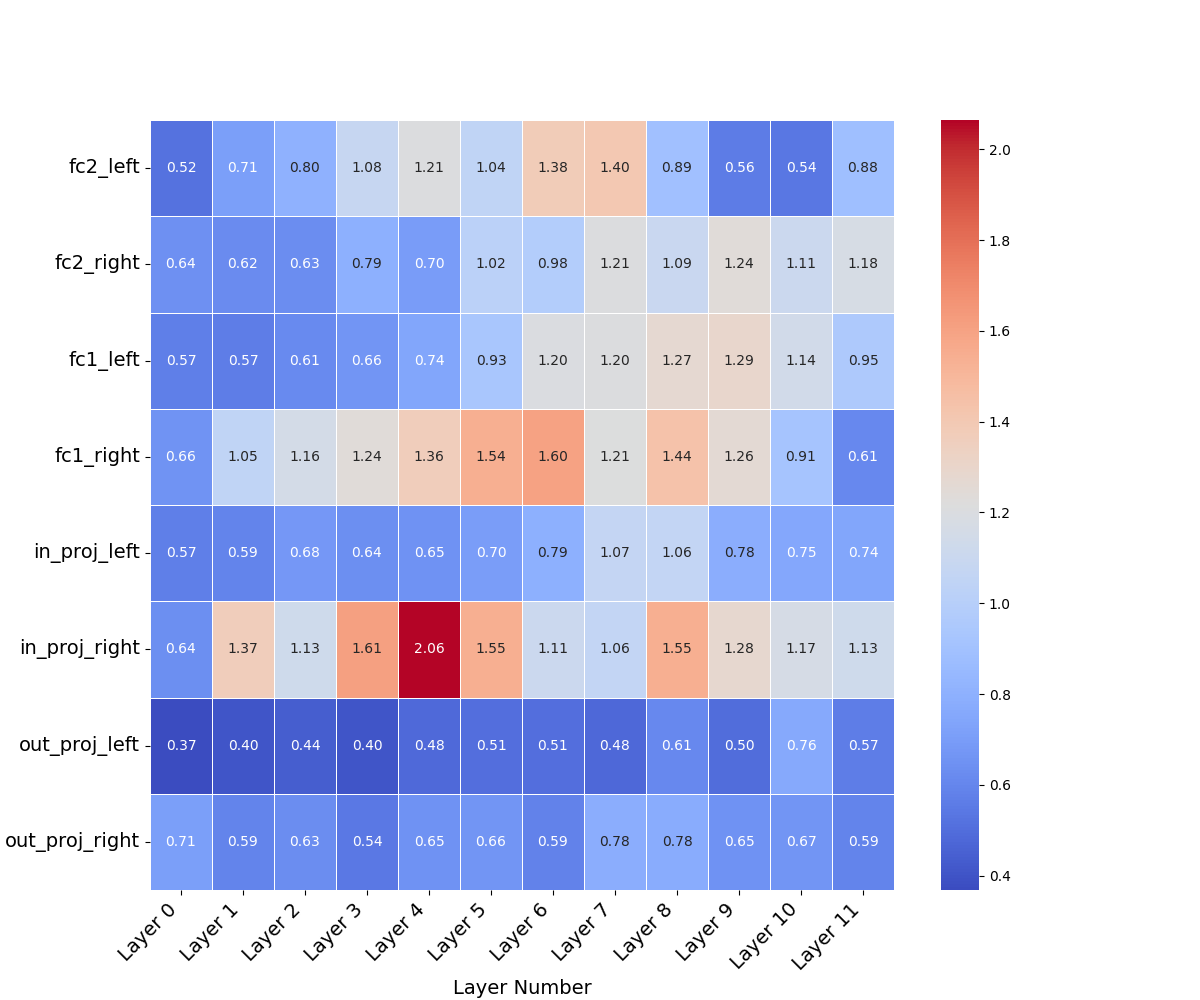}}
    \caption{Distribution of noise multipliers during the training of \adanoise{} on the SST-2 dataset. The X-axis represents the 12 layers of the Roberta-base model, while the Y-axis denotes the PEFT weights. The color gradient indicates the varying amounts of noise applied.}
    \label{fig:noise_multiplier}
\end{figure*}

The performance of \adanoise{} in comparison to past DP methods and regular training is given in \cref{tab:performance_metrics}.
Introducing privacy protection inevitably leads to performance degradation. 
Nevertheless, we observe that \adanoise{} consistently outperforms traditional DP-SGD full fine-tuning and PEFT fine-tuning, demonstrating the benefit of using \adanoise{}.
For instance, \adanoise{} achieves a performance improvement of 1.4\% for RoBERTa-large on the QQP task in the Adapter setting, and 1.5\% for RoBERTa-base on the QNLI task in the LoRA setting. For full fine-tuning, \adanoise{} also poses a performance gain of up to 1.7\% compared to the conventional DP-SGD. 
The improvement is consistent across all the tasks and settings. In our additional experiments, we found that \adanoise{} only introduces less than 1\% more GPU memories and 5\% more training time compared to the original DP method, yet achieves better utility. The performance differences between \adanoise{} and DP are all statistically significant with p < 0.05 under one-tailed paired t-test. Finally, in order to better understand how \adanoise{} differs from past DP techniques, we examine the detailed noise distribution breakdown in Figure~\ref{fig:noise_multiplier}.

\subsection{Exposure Risks Results}
Our exposure experiments seek to assess the risk of privacy leakage empirically, particularly focusing on sensitive information such as person names and telephone numbers.
Such an experiment is important as concerns have previously been raised on whether DP guarantees adhere to the allocated privacy budget \citep{steinke2024privacy}.
This discrepancy can occur due to various factors, ranging from theoretical assumptions not holding in practice to statistical variations and implementation bugs. 


\begin{figure}[t]
    \centering
    \includegraphics[width=0.38\textwidth,height=0.15\textheight]{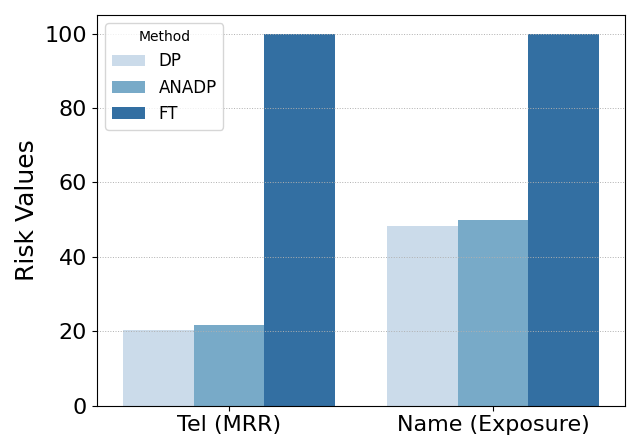}
    \caption{Privacy leakage risks for telephone number and person name using DP, \adanoise{} and non-DP full fine-tuning.}
    \label{fig:Privacy_Guarantees}
\end{figure}


\cref{fig:Privacy_Guarantees} shows that \adanoise{} maintains the same level of privacy protection as the conventional DP methods on the Enron email dataset~\cite{klimt2004enron}, without statistically significant difference between them. The exposure risk values show a substantial reduction compared to those of the non-DP model, demonstrating its effectiveness in mitigating privacy leakage risks.
Overall, with the same privacy protection capability as conventional DP, \adanoise{} consistently improves the performance of the latter in the benchmark tasks described above, benefiting from considering the different contributions of parameters.

\subsection{\adanoise{} Noise Distribution}\label{sec:appendix_noise}

Figure~\ref{fig:noise_multiplier} shows the detailed distribution of noise multipliers applied via \adanoise{} when tuning RoBERTa-base on the SST-2 task. 
\adanoise{} demonstrates a strategic pattern in noise allocation, consistently applying lower noise levels to more critical parameters. 
Notably, the lower and final layers of the model often receive reduced noise. This could suggest that initial layers, responsible for capturing basic linguistic features, and final layers, which fine-tune these features into task-specific outputs, are deemed more sensitive to noise disruption.
This pattern supports the hypothesis that maintaining the integrity of these parameters is crucial for preserving the model's overall performance.


In contrast, \adanoise{} strategically assigns higher noise levels to the middle layers of the transformer model. 
This allocation aligns with the findings of \citet{meng2023locating} which concluded that factual knowledge is predominantly stored in the middle layers of the feed-forward network.
By introducing more noise to these layers, \adanoise{} effectively obfuscates sensitive factual associations, thereby enhancing privacy protection without compromising the model's ability to learn and perform on specific tasks.
This targeted noise allocation ensures that while the privacy of stored knowledge is robustly safeguarded, the overall performance of the model remains optimized.

\section{Conclusion}
This paper introduces \adanoise{}, a novel approach to integrating DP with language model fine-tuning in both the full fine-tuning and PEFT settings and dynamically adjusting the noise added to the gradients, based on measuring model parameters' importance.
We demonstrated that under the same privacy budget, \adanoise{} consistently outperforms the standard DP-SGD training on different benchmark datasets.
While performance degradation remains between our method and non-DP training, we achieved consistent reduction of the gap, in both the fine-tuning and PEFT settings.
Our additional exposure risk analysis shows that \adanoise{} provides privacy protection comparable to the standard DP-SGD training. We hope this work enables better deployment of privacy-preserving language models and encourages future research on adaptive DP for language model training.

\section{Limitations}
While \adanoise{} offers consistent improvements, there are certain limitations that present opportunities for future work. 
First, although our method effectively identifies important parameters for downstream tasks and allocates noise accordingly, it does not explicitly distinguish whether these parameters are also privacy-sensitive. Identifying privacy-related parameters during the training process could be a crucial research problem. 
Moreover, developing an automated method to normalize noise would significantly streamline the application of \adanoise{}.

\section*{Acknowledgement}

This research was funded in part by the Faculty Research Awards of J.P. Morgan AI Research. The authors are solely responsible for the contents of the paper and the opinions expressed in this publication do not reflect those of the funding agencies.
\\
\section*{Disclaimer}

This paper was prepared for informational purposes in part by the Artificial Intelligence Research group of JPMorgan Chase \& Co. and its affiliates (``JP Morgan''), and is not a product of the Research Department of JP Morgan. JP Morgan makes no representation and warranty whatsoever and disclaims all liability, for the completeness, accuracy or reliability of the information contained herein. This document is not intended as investment research or investment advice, or a recommendation, offer or solicitation for the purchase or sale of any security, financial instrument, financial product or service, or to be used in any way for evaluating the merits of participating in any transaction, and shall not constitute a solicitation under any jurisdiction or to any person, if such solicitation under such jurisdiction or to such person would be unlawful.

\bibliography{custom}

\clearpage
\appendix
\section{Appendix A}
\label{sec:appendix_experiment}

\paragraph{Training Details.} 
Following prior work in model privacy, we conduct our training using RoBERTa (on both the \textit{base} and \textit{large} version) \cite{liu2019roberta}. 
RoBERTa-base has 12 transformer layers, a hidden state size of 768, and a feedforward network (FFN) with an internal hidden size of 3072. 
RoBERTa-large is configured with 24 transformer layers, enhancing its complexity. In LoRA fine-tuning, we followed \citet{yu2021differentially} where we incorporated bottleneck branches in both the attention layers and the feedforward layers.
This approach differs slightly from the method used by \citet{hu2021lora}, who only added bottleneck branches to the query and values matrices within the attention layers.
For the two types of PEFT methods, we choose the same rank 16 for all the experiments.
For DP experiments, we use $\epsilon$=8, C=10, $\delta$=1e-5 for SST-2, QNLI and $\delta$=1e-6 for MNLI, QQP dataset.
We run 50 epochs on all datasets and report the best validation accuracy. All experiments were conducted using NVIDIA A100 GPUs.

\begin{figure}[ht]
    \centering
    \resizebox{\linewidth}{!}{\includegraphics[width=\linewidth]{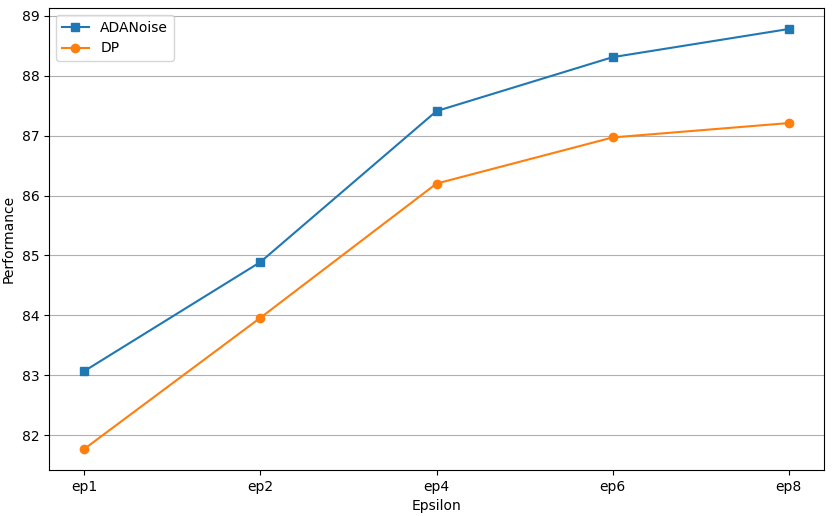}}
    \caption{Performance of \adanoise{} under different privacy budget on QNLI dataset.}
    \label{fig:epsilons}
\end{figure}

We further extend our analysis by examining the performance of \adanoise{} under different privacy budgets, as illustrated in figure \ref{fig:epsilons}.
The trends demonstrate that \adanoise{} consistently outperforms traditional DP under all scenarios, even when the privacy budget is set to a relatively low value ($\epsilon = 1$). 
This difference becomes more pronounced as \(\epsilon\) increases, with \adanoise{} reaching 88.78 at \(\epsilon = 8\) compared to DP's 87.21.
This consistent outperformance highlights the effectiveness of \adanoise{}.

\begin{table}[htb]
\centering
\small
\caption{Comparison of \adanoise and DP with BERT-base.}
\label{tab:bert_results}
\begin{tabular}{cccc}
\hline
\textbf{Task} & \textbf{Method} & \textbf{BERT-base} \\ \hline
\multirow{2}{*}{SST-2} & ANADP & 88.18 \\ 
                       & DP    & 87.27 \\ \hline
\multirow{2}{*}{QNLI}  & ANADP & 86.61 \\ 
                       & DP    & 86.03 \\ \hline
\multirow{2}{*}{MNLI}  & ANADP & 79.46 \\ 
                       & DP    & 78.85 \\ \hline
\multirow{2}{*}{QQP}   & ANADP & 84.97 \\ 
                       & DP    & 84.72 \\ \hline
\end{tabular}
\end{table}

 we have included additional experimental results from BERT in the table below. These results demonstrate the effectiveness and versatility of ANADP across various models, reinforcing its generalizability. It is also worth noting that, in the current studies on DP-based models, it is not common to use larger generative models, and most packages do not support such models. We follow the same setup to make our work comparable to the existing work. We leave the investigation on larger generative models as future work.

\end{document}